\documentclass[sigconf]{acmart}
\usepackage{amsmath}
\usepackage{multirow}
\AtBeginDocument{%
  \providecommand\BibTeX{{%
    \normalfont B\kern-0.5em{\scshape i\kern-0.25em b}\kern-0.8em\TeX}}}

\setcopyright{none}
\copyrightyear{}
\acmYear{}
\acmDOI{}

\acmConference{}
\acmBooktitle{}
\acmPrice{}
\acmISBN{}

\acmSubmissionID{816}

\begin{document}

\title{Cross-Level Cross-Scale Cross-Attention Network for Point Cloud Representation}

\author{Xian-Feng Han}
\email{xianfenghan@swu.edu.cn}
\orcid{0000-0002-4869-4537}
\affiliation{%
  \institution{College of Computer and Information Science, Southwest University}
  \streetaddress{No.2 Tiansheng Road}
  \city{Beibei District}
  \state{Chongqing}
  \country{China}
  \postcode{400712}
}

\author{Zhang-Yue He}
\email{hezhangyue@email.swu.edu.cn}
\affiliation{%
  \institution{College of Computer and Information Science, Southwest University}
  \streetaddress{No.2 Tiansheng Road}
  \city{Beibei District}
  \state{Chongqing}
  \country{China}
 \postcode{400712}
}

\author{Jia Chen}
\email{gqxiao@swu.edu.cn}
\affiliation{%
  \institution{College of Computer and Information Science, Southwest University}
  \streetaddress{No.2 Tiansheng Road}
  \city{Beibei District}
  \state{Chongqing}
  \country{China}
  \postcode{400712}
}

\author{Guo-Qiang Xiao}
\email{gqxiao@swu.edu.cn}
\affiliation{%
  \institution{College of Computer and Information Science, Southwest University}
  \streetaddress{No.2 Tiansheng Road}
  \city{Beibei District}
\state{Chongqing}
  \country{China}
  \postcode{400712}
}


\begin{abstract}
Self-attention mechanism recently achieves impressive advancement in Natural Language Processing (NLP) and Image Processing domains. And its permutation invariance property makes it ideally suitable for point cloud processing. Inspired by this remarkable success, we propose an end-to-end architecture, dubbed Cross-Level Cross-Scale Cross-Attention Network (CLCSCANet), for point cloud representation learning. First, a point-wise feature pyramid module is introduced to hierarchically extract features from different scales or resolutions. Then a cross-level cross-attention is designed to model long-range inter-level and intra-level dependencies. Finally, we develop a cross-scale cross-attention module to capture interactions between-and-within scales for representation enhancement. Compared with state-of-the-art approaches, our network can obtain competitive performance on challenging 3D object classification, point cloud segmentation tasks via comprehensive experimental evaluation.
\end{abstract}

\ccsdesc{3D point cloud analysis}
\ccsdesc{Computing methodologies~Neural Networks}

\keywords{Point Cloud, Self Attention, Classification, Segmentation}

\begin{teaserfigure}
  \includegraphics[width=\textwidth]{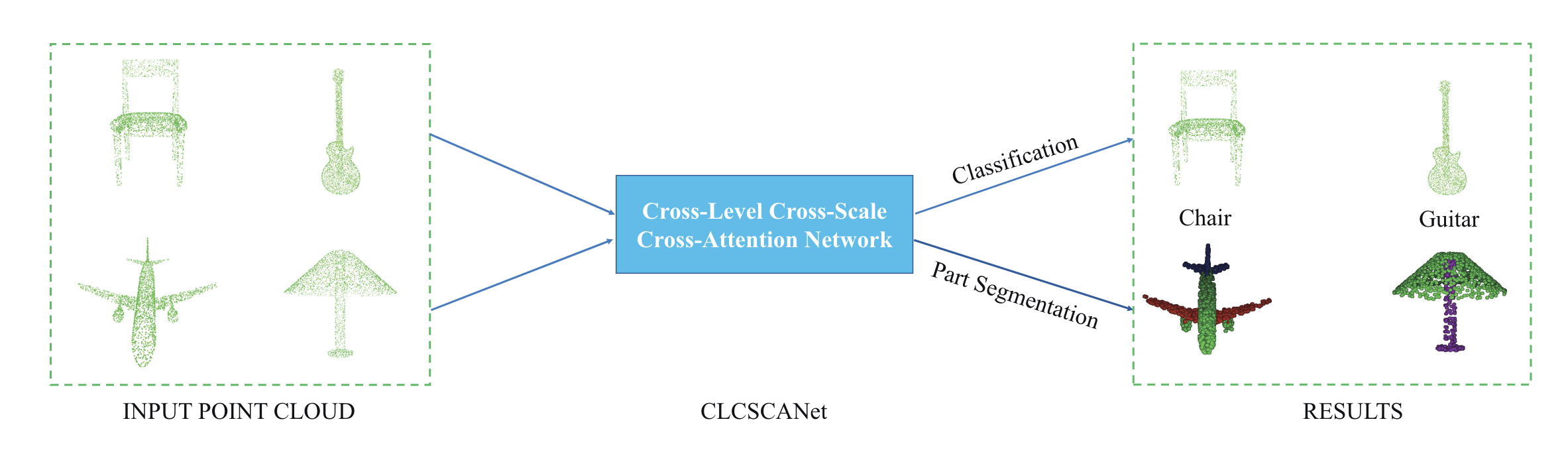}
  \caption{Applications of our Cross-Level Cross-Scale Cross-Attention Network.}
  \label{fig:teaser}
\end{teaserfigure}

\maketitle

\section{Introduction}
The rapid development of 3D acquisition devices makes point cloud become a preferred 3D geometric representation for arbitrarily-shaped objects in real word \cite{te2018rgcnn}. And thanks to its flexibility, analyzing and understanding 3D point clouds, therefore, begin to serve as a fundamental processing demanded by a number of applications, such as robotics \cite{nezhadarya2020adaptive}, autonomous driving \cite{guo2020deep} and VR/AR \cite{gojcic2020learning}.  

Recently, deep learning techniques has revolutionized 2D computer vision with powerful discrete convolution as essential component for impressive representation learning on regular grid. Their outstanding performance inspires more and more attention paid to deep learning on 3D point clouds. However, since 3D point cloud can be considered as an unordered collection of points with irregular structure, it is not always reasonable to directly transfer deep learning techniques for image processing to point cloud based tasks. 

In order to sovle this challenge, several attempts have been made to transform the unstructured 3D point clouds into regular girds, such as voxel grid \cite{maturana2015voxnet}, multi-view images \cite{su2015multi}, so that standard 3D CNNs or 2D CNNs can be applied for feature learning. Nevertheless, these approaches usually requires high memory footprint, computational consumption, and may suffer from information loss during transformation. More recently, deep neural networks directly manipulating raw point clouds have been specifically developed. As a pioneering work, PointNet \cite{qi2017pointnet} encodes each individual point identically, and then accumulates point features into a global vector via a symmetric function. PointNet++ \cite{qi2017pointnet++} improves PointNet by taking into account local information of 3D point cloud. However, these point-wise methods essentially treat points independently \cite{han2020point2node} and fail to capture their long-range correlation relationship \cite{wang2018dynamic}.

To better leverage point dependencies between and in different levels and scales, we propose a novel architecture, named Cross-Level Cross-Scale Cross-Attention network (CLCSCANet), for effective point feature learning, which mainly contains three components. Specifically, we first design a Point Feature Pyramid (PFP) module to hierarchically extract point-wise features with different receptive fields. Then, a Cross-Level Cross-Attention model is developed to investigate inter-level and intra-level correlations for geometric and semantic information aggregation. Finally, a Cross-Level Cross-Attention module is designed to learn inter-scale and intra-scale interactions to further improve the discriminative ability of feature representations.

We conduct extensive quantitative and qualitative evaluation on ModelNet40 \cite{wu20153d} for 3D object classification, ShapNet \cite{yi2016scalable} for part segmentation. Experimental results demonstrate the effectiveness and capability of better feature representation learning of our CLCSCANet with achievement of comparable performance. 


Summarily, the major contributions of this work are concluded as follows:
\begin{itemize}
   \item Three well-designed modules, namely Point-wise Feature Pyramid, Cross-Level Cross-Attention, Cross-Scale Cross-Attention, are developed to aggregate features across different levels and scales, while modeling their long-range correlation dependencies. 
   
   \item Based on these models, we build an end-to-end Cross-Level Cross-Scale Cross-Attention Network (CLCSCANet) architecture taking point clouds as input for highly geometric and semantic representation learning.
   
   \item We perform extensive experiments on publicly available benchmark datasets across three tasks (i.e. classification and segmentation) and verify the effectiveness of our CLCSCANet framework which can obtain  competitive performance compared with the state-of-the-arts.
\end{itemize}

\section{Related Work}

\subsection{Deep Learning on Point Clouds}
The recent breakthroughs from deep learning in 2D computer vision domain has been shifting the research trend gradually to its application in 3D point clouds. However, this task is quite challenging, since standard CNNs are only suitable for data with regular structure. 

\noindent\textbf{Grid-based Methods.} In order to solve such problem, early attempts mainly focus on transforming irregular point clouds into an intermediate grid structure. View-based methods \cite{su2015multi}\cite{feng2018gvcnn} intuitively aims to turn the 3D problems into 2D problems by projecting the 3D point clouds to a set of images from different views. Then, conventional convolutional neural networks in 2D images can be used to perform feature extraction. Finally, these features are aggregated back into 3D space. Although remarkable performance has been achieved in classification task, it is nontrivial to perform segmentation task due to the loss of intrinsic geometric relationship of 3D points during projection \cite{han2020point2node}. On the other hand, the choice of images including the number of views also seriously impact on performance \cite{zhao2019pointweb}. 
 
Alternatively, voxel-based approaches attempt to assign the scattered 3D points to a regular 3D grid structure, such as occupancy voxel grid \cite{maturana2015voxnet}, that can be processed by standard 3D convolutional neural networks. However, the cubic growth of memory requirements and computational demanding limit their applicable to high-resolution voxelization \cite{lei2019octree}, and also constrains the expressiveness and efficiency of voxelized representation of point clouds \cite{yang2020pbp}. To alleviate this problem, the octree-based \cite{lei2019octree} and kd-tree-based \cite{klokov2017escape} models are proposed to refine the performance \cite{lin2020convolution}. For example, OctNet \cite{riegler2017octnet} improves the resolution up to $256 \times 256 \times 256$. Nevertheless, the grid-based representation still suffers from quantization artifacts and fine-grained information loss.  

\noindent\textbf{Point-based Methods.}The appearance of PointNet \cite{qi2017pointnet} leads to a new trend of deep learning on point clouds. This network directly takes coordinates of points as input and uses shared MLPs to independently learn point-wise features, which are then aggregated into a global representation via the channel-wise max-pooling. A major limitation is that PointNet lacks ability to acquiring local information of point clouds. To remedy this, PointNet++ \cite{qi2017pointnet++} extends PointNet with a hierarchical structure to understand local features using farthest ponit sampling and a geometric grouping layer \cite{zhao2019pointweb}\cite{xie2018attentional}. 
Some latest works focus on generalizing convolution operator for grid-structured data to handle unstructured point clouds. And these methods usually define spatial kernels by exploring the spatial localization information in a point clouds \cite{thomas2019kpconv}. Kernel Point Convolution uses a small collection of kernel points for local feature learning. PointConv \cite{wu2019pointconv} can be considered as Monte Carlo estimation of 3D continuous convolution, which uses MLP and a density function to estimate and re-weight the learned weight function. \cite{lin2020fpconv} designs a surface-style convolution, named FPConv, for point cloud analysis. \cite{engelmann2020dilated} proposes Dilated Point Convolutions to increase the receptive field size.

Unlike these methods, we pay attention not only to acquiring information in local regions, but also to capturing long-range context dependencies among points in point clouds.
\subsection{Self-Attention}
Attention mechanism originally from natural language processing domain \cite{bahdanau2014neural}\cite{lin2017structured}, nowadays, has been attracting more and more attention in the field of computer vision, since it has the ability of capturing global contextual information by modeling much broader range dependencies between the components of the input sequence. For example, Fu et al. \cite{fu2019dual} developed position attention and channel attention modules based on self-attention mechanism for scene segmentation. On the other hand, the usage of matrix multiplication and summation in self-attention makes it invariant to permutation, which is an ideal choice for point cloud processing. 

Inspired by this remarkable property, we design the cross-level cross-attention and cross-scale cross-attention modules with the fundamental idea of self-attention for discriminative representation learning.

\section{Cross-Level Cross-Scale Cross-Attention Network}
\subsection{Overview}
\begin{figure*}[h]
  \centering
  \includegraphics[width=7in]{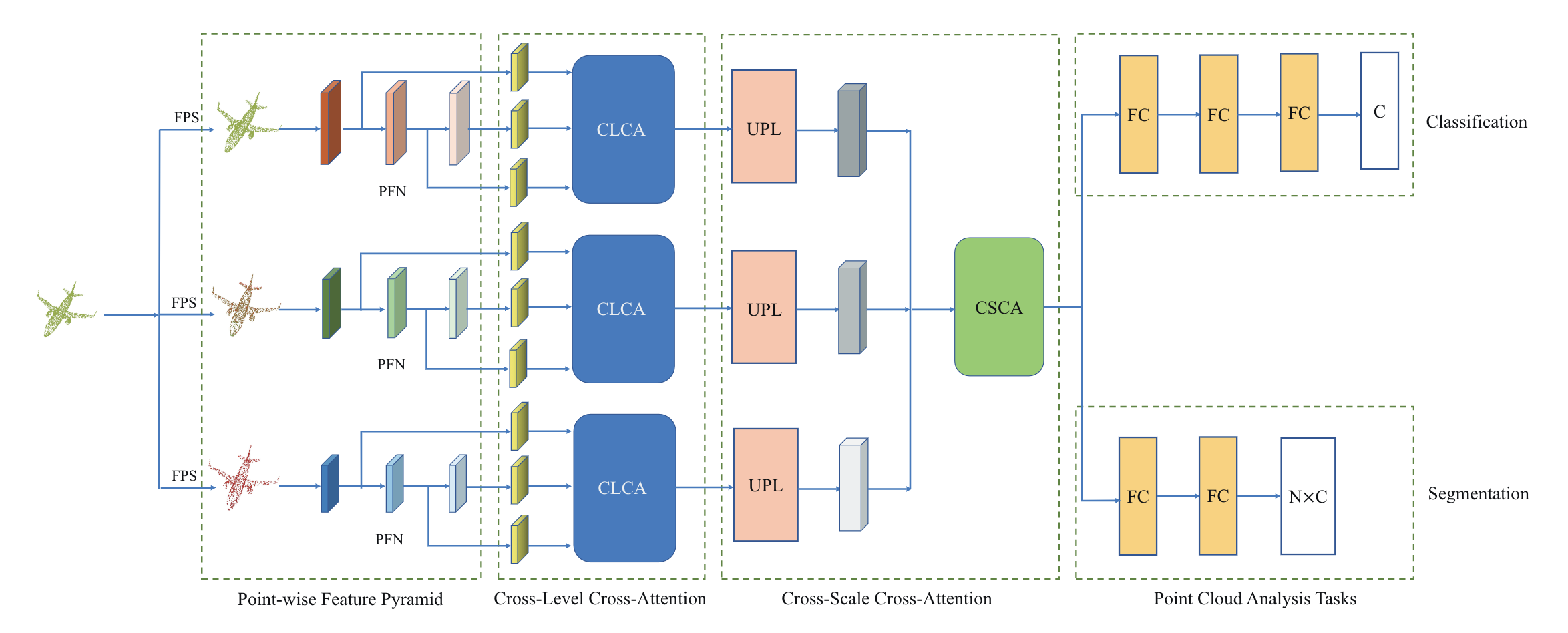}
  \caption{Overall Architecture of proposed Cross-Level Cross-Scale Cross-Attention Network for Point Cloud Analysis. The network mainly consists of three core modules, namely Point-wise Feature Pyramid, Cross-Level Cross-Attention and Cross-Scale Cross-Attention. C and D denote the number of classes for Classification and Segmentation tasks.}
  \label{fig_framework}
\end{figure*}

Figure \ref{fig_framework} illustrates the overall architecture of our proposed end-to-end Cross-Level Cross-Scale Cross-Attention Network (CLCSCANet). Our goal is to learn a projection function $f: P \rightarrow R^{F}$ for various point cloud based tasks including 3D classification and segmentation. The network directly undertakes 3D point cloud wiht $N$ points $P = \{p_{i} \in R^{3+a}, i = 1, 2, ..., N\} $ as input, where $3+a$ represents 3D coordinates $(x_{i}, y_{i}, z_{i})$ and additional attributes, such as color, surface normal. First, Farthest Point Sampling algorithm is initially used to obtain three sub point clouds with different resolutions, each of which will be fed into the corresponding path of our Point Feature Pyramid module (Section \ref{pyramid}) to construct hierarchical features. Then, we explore intra-level and inter-level point feature relationships by building a Cross-Level Cross-Attention (Section \ref{crosslevel}) model for aggregating geometric and semantic information simultaneously. Finally, a Cross-Scale Cross-Attention module (Section \ref{crossscale}) is defined to fully investigate the correlation among points within and from different scales for feature representation enhancement. In the following sections, we will elaborate these three modules.
\subsection{Point-wise Feature Pyramid}
Empirically, the receptive field size is directly related to contextual information learning (larger receptive field usually means description of much broader context ), which makes significant contribution to the performance of 3D point cloud processing \cite{engelmann2020dilated}. Hence, in order to extract hierarchical features to deal with objects with diverse scales, we develop a Point-Wise Feature Pyramid (PFP) module, which is composed of three individual branches (from top to bottom, the resolution is decreasing while the receptive field size is increasing). 

Specifically, as illustrated in Figure \ref{fig_framework}, we first perform Furthest Point Sampling (FPS) algorithms \cite{qi2017pointnet++} on initial input point cloud $P$ to obtain three subsets of points with different resolutions $\{P_{1} \in R^{N1\times (3+a)}) = FPS(P, N_{1}), P_{2} \in R^{N2\times (3+a)}= FPS(P, N_{2}), P_{3} \in R^{N3\times (3+a)}=FPS(P, N_{3})\}$, since FPS can generate limited centroids ensuring better coverage of point cloud and maintain the original shape \cite{lin2020fpconv}. Then, $K$ points around each centroid that are determined via ball query with a given radius are grouped to form a local region for capturing local feature. Finally, the three independent paths in PFP module apply their corresponding mapping functions $f_{scale}^{i}: R^{N_{i} \times (3+a)} \rightarrow R^{N_{i} \times C}, i = 1, 2, 3$ to these three downsampled point clouds to extract three individual latent representations. In our work, we define each path by stacking multiple shared Multi-Layer Perception (MLP) modules. 
\label{pyramid}

\subsection{Cross-Level Cross-Attention Module}
\label{crosslevel}
\begin{figure*}[h]
  \centering
  \includegraphics[width=7in]{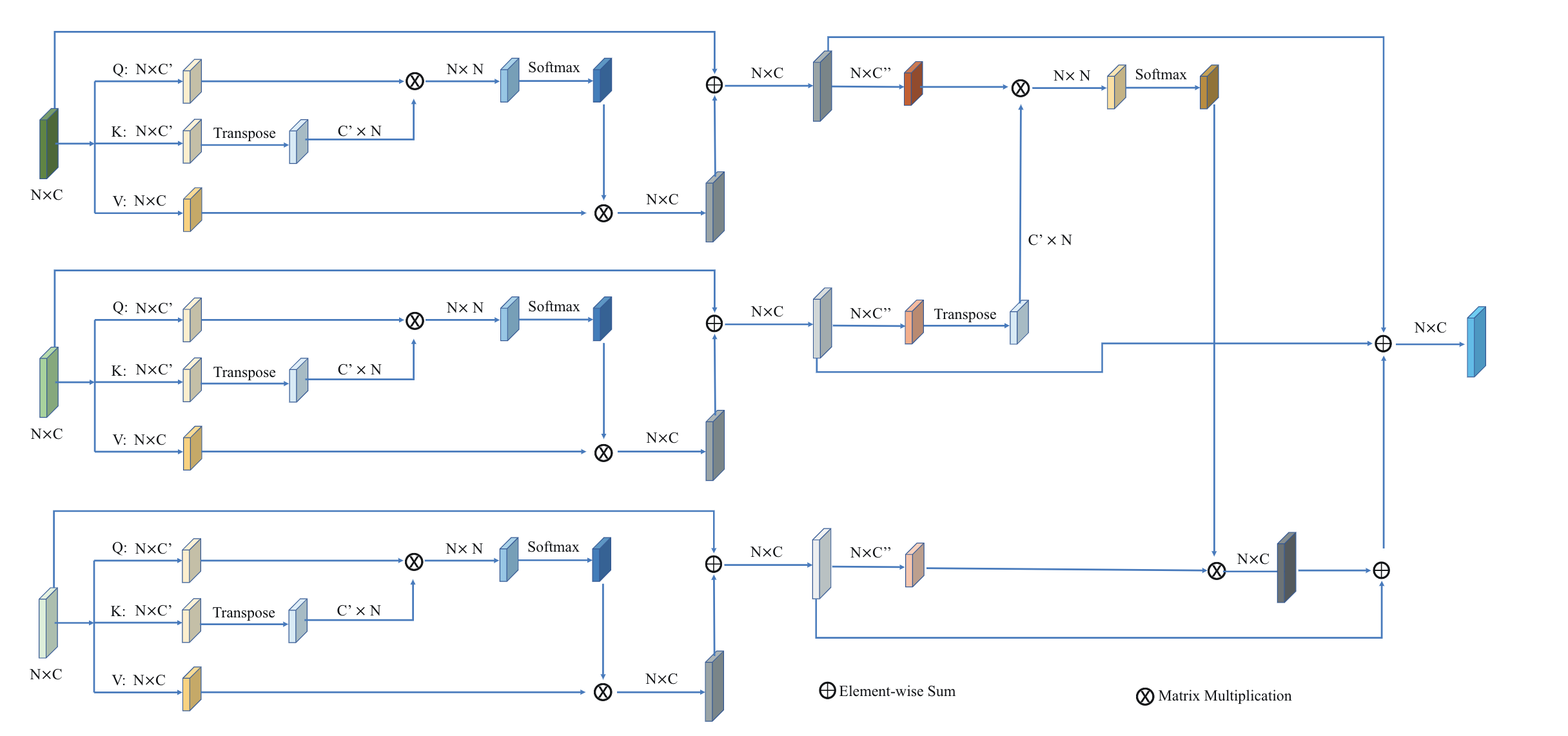}
  \caption{Cross-Level Cross-Attention Architecture for inter-level and intra-level dependencies learning.}
  \label{fig_clsa}
\end{figure*}

The spatial dependencies among points in the same level and among different levels can provide both semantically-rich and geometrically-rich information. In order to better model intra-level and inter-level correlation and make full use features from different levels, we formulate the Cross-Level Cross-Attention (CLCA) model for learning more comprehensive feature representation. The overall architecture is shown in Figure \ref{fig_clsa}.

\noindent \textbf{Intra-Level Correlation.} For each path $i = 1, 2, 3$, we separately extract low-level, mid-level and high-level feature maps, denoting as $\mathcal{F}_{low}^{i}$, $\mathcal{F}_{mid}^{i}$, $\mathcal{F}_{high}^{i}$, from corresponding shared MLP layers. Here, we will take the high-level feature map $\mathcal{F}_{high}^{i}$ as example for simplified description. First, we apply poinwise feature transformations to $\mathcal{F}_{high}^{i}$ to calculate the query, key and value matrices, which are defined as:
\begin{equation}
    Q_{high}^{i} = \phi(\mathcal{F}_{high}^{i}) = \mathcal{F}_{high}^{i}W_{high-i}^{q}
\end{equation}
\begin{equation}
    K_{high}^{i} = \psi(\mathcal{F}_{high}^{i}) = \mathcal{F}_{high}^{i}W_{high-i}^{k}
\end{equation}
\begin{equation}
    V_{high}^{i} = \beta(\mathcal{F}_{high}^{i}) = \mathcal{F}_{high}^{i}W_{high-i}^{v}
\end{equation}

Then, the formulation for point self correlations $F_{i}$ is modeled as follows:
\begin{equation}
    SC^{i}_{high} = \sigma(Q_{high}^{i} \cdot (K_{high}^{i})^T/\sqrt{C‘})V_{high}^{i} + \mathcal{F}_{high}^{i}
\end{equation}
where $\phi(\cdot)$, $\psi(\cdot)$, $\beta(\cdot)$ are linear transformations in our work, and $W_{mid-i}^{q} \in R^{C \times C'}$, $W_{mid-i}^{k}\in R^{C \times C'}$, $W_{mid-i}^{v} \in R^{C \times C} $ are corresponding learnable weight parameters matrices. We set $C' = C/4$ for computational efficiency. $\sigma(\cdot)$ is a normalization function, and softmax is chosen in our work. Simultaneously, the same procedures are applied to low-level and mid-level features to obtain the corresponding outputs$SC^{i}_{low}$ and $SC^{i}_{mid}$.

\noindent \textbf{Inter-Level Correlation.}To further explore correlations across different levels for aggregating more discriminate point-wise features, we introduce the cross-attention for level interaction, which can be formulated as follows:
\begin{equation}
       F_{CLCA}^{i} = \mathcal{AT}(SC^{i}_{low}, SC^{i}_{mid}, SC^{i}_{high}) + SC^{i}_{low} + SC^{i}_{mid} + SC^{i}_{high}
\end{equation}
where,
\begin{align}
    \mathcal{AT}(\cdot) &= \sigma(f_{1}(SC^{i}_{low}) \cdot (f_{2}(SC^{i}_{mid})^T) \cdot f_{3}(SC^{i}_{high})\\
    & = \delta((SC^{i}_{low}W_{clca-i}^{1}) \cdot (SC^{i}_{mid}W_{clca-i}^{2})^T) \cdot (SC^{i}_{high}W_{clca-i}^{3})
\end{align}
Here, $W_{clca-i}^{1} \in R^{C \times C/4}$, $W_{clca-i}^{2} \in R^{C \times C/4}$ and $W_{clca-i}^{3} \in R^{C \times C}$ are linear projection parameters matrices. 

Lastly, the final output of our Cross-Level Cross-Attention module contains three individual representations $F_{CLCA}^{i} \in R^{N_{i}\times C}, i = 1, 2, 3$, each having the same scales as the corresponding $P_{i}$. And the usage of multi-level information is actually of important in point cloud understanding.

\subsection{Cross-Scale Cross-Attention Module}
\label{crossscale}
In PFP module, we see that the top branch has the highest resolution but lowest receptive field, while the bottom with lowest resolution and largest receptive field. Empirically, point-wise feature at different scales generally contains different semantic or contextual information. Therefore, to fully exploit long-range self correlation within the same scale and aggregate dependencies across different scale correlation learning, we formulate Cross-Scale Cross-Attention module for enhancing the discriminative power of representation. 

\noindent \textbf{Up-Sampling Layer.} Specifically, to perform point-wise prediction task conveniently, such as segmentation, we should first propagate point-wise feature maps  $F_{CLCA}^{i}$ from CLCA up to the original resolution of input point cloud through up-sampling layer that mainly uses $K$ nearest neighbors (KNN) interpolation and shared MLPs. 
\begin{equation}
    F_{cs-in}^{i} = UP(F_{CLCA}^{i}) = MLP(KNN-Interpo(F_{CLCA}^{i}, P))
\end{equation}
Where $F_{CSCA-input}^{i} \in R^{N \times D}$ represents the interpolated point-wise feature maps from the $i$th scale ($i = 1, 2, 3$).

\noindent \textbf{Intra-Scale Interaction.}Then, based on the fundamental idea of self-attention mechanism, relationship among points of $i$th scale can be constructed as:

\begin{equation}
    SC_{scale}^{i} = \sigma((F_{cs-in}^{i}W_{SC}^{q})\cdot(F_{cs-in}^{i}W_{SC}^{k})^{T}/sqrt(D'))\cdot(F_{cs-in}^{i}W_{SC}^{v}) + F_{cs-in}^{i}
\end{equation}
where $W_{SC}^{q} \in R^{D \times D'}$, $W_{SC}^{k} \in R^{D \times D'}$ and $W_{SC}^{v} \in R^{D \times D}$ are the weight parameters of three fully-connected layers. And $D'$ is set to $D / 4$.

\noindent \textbf{Inter-Scale Interaction.} Similar to Cross-Level Cross-Attention model, we also consider the cross scale attention to fuse multi-scale information. And this interaction model is formalized as:
\begin{equation}
    F_{CSCA} = \mathcal{AT'}(SC_{scale}^{1}, SC_{scale}^{2}, SC_{scale}^{3})+SC_{scale}^{1}+SC_{scale}^{2}+SC_{scale}^{3}
\end{equation}
Here, 
\begin{equation}
    \begin{split}
        AT' = \sigma((SC_{scale}^{1}W^{1})\cdot(SC_{scale}^{2}W^{2})^{T})\cdot((SC_{scale}^{3}W^{3})) \\
        W^{1} \in R^{D \times D'}, W^{2} \in R^{D \times D'}, W^{3} \in R^{D \times D}
    \end{split}
\end{equation}

The Cross-Scale Cross-Attention module finally outputs semantically meaningful point-wise feature representation. 

\subsection{Architecture and Supervision}
From Figure \ref{fig_framework}, it can be clearly seen that these three modules depicted in the previous subsections, together with fully connected layers, constitute the overall architecture of our proposed CLCSCANet, which is simple and easy to implement..
The detailed configuration of 3D object classification and point cloud segmentation networks are provided in Table \ref{}. During training, cross-entropy is employed to supervise the learning process of our model in an end-to-end manner. 
\begin{equation}
   CE(G,P)= \left\{
     \begin{aligned}
         -\sum_{i = 1}^{C}G_{i}log(P_{i}) & \qquad for \; classification\\
        -\sum_{n = 1}^{N}\sum_{i = 1}^{C}G_{n,i}log(P_{n,i}) & \qquad for \; segmentation\\
\end{aligned}
\right.
\end{equation}
where G and P represent the ground truth and prediction, respectively. N refers to the number of input points. C denotes the number of classes.

\subsection{Network Configurations}
The detailed configurations of network for classification and segmentation tasks are summarized in Table \ref{tab:networkcon} 

\begin{table*}[]
\caption{Network configurations for classification and segmentation. NN($r, K$) denotes the radius of ball query and the number of points to group for local information extraction. MLP($C_{in}, C_{out}$) represents the Multiple Layer Perceptron, taking $C_{in}$ input features and generating $C_{out}$ features. CLCA($C_{in}$) and CSCA($C_{in}$) denotes cross-level cross-attention and cross-scale cross-attention operations for modeling interactions. FC($C_{in}, C_{out}$) means fully connected layer. $C$ is the number of categories of corresponding dataset.}

\label{tab:networkcon}
\scalebox{0.9}{
\begin{tabular}{cccccccccc}
\hline
\multicolumn{1}{l}{Tasks}       & \multicolumn{1}{l}{Scales} & \multicolumn{1}{l}{Resolution} & \multicolumn{1}{l}{Layer1}                                                                          & \multicolumn{1}{l}{Layer2}                                                             & \multicolumn{1}{l}{Layer3}                                                              & \multicolumn{1}{l}{Layer4} & \multicolumn{1}{l}{Layer5}                                                                       & Layer6                     & Layer7                                                                                             \\
\hline
\multirow{3}{*}{Classification} & Path1                      & 512                            & \begin{tabular}[c]{@{}c@{}}NN(r=0.2, K=16)\\ MLP(3, 64)\\ MLP(64, 128)\\ MLP(128, 128)\end{tabular} & \begin{tabular}[c]{@{}c@{}}NN(r=0.4, K=32)\\ MLP(256,128)\\ MLP(128, 128)\end{tabular} & \begin{tabular}[c]{@{}c@{}}NN(r=0.8, K=64)\\ MLP(256,128)\\ MLP(128, 128)\end{tabular}  & CLCA(256)                  & -                                                                                                & \multirow{3}{*}{CSCA(512)} & \multirow{3}{*}{\begin{tabular}[c]{@{}l@{}}FC(1024, 512)\\ FC(512, 256)\\ FC(256, C)\end{tabular}} \\
                                & Path2                      & 256                            & \begin{tabular}[c]{@{}c@{}}NN(r=0.2, K=16)\\ MLP(3, 64)\\ MLP(64, 128)\\ MLP(128, 128)\end{tabular} & \begin{tabular}[c]{@{}c@{}}NN(r=0.4, K=32)\\ MLP(256,128)\\ MLP(128, 128)\end{tabular} & \begin{tabular}[c]{@{}c@{}}NN(r=0.8, K=64)\\ MLP(256,128)\\ MLP(128, 128)\end{tabular}  & CLCA(256)                  & -                                                                                                &                            &                                                                                                    \\
                                & Path3                      & 128                            & \begin{tabular}[c]{@{}c@{}}NN(r=0.2, K=16)\\ MLP(3, 64)\\ MLP(64, 128)\\ MLP(128, 128)\end{tabular} & \begin{tabular}[c]{@{}c@{}}NN(r=0.4, K=32)\\ MLP(256,128)\\ MLP(128, 128)\end{tabular} & \begin{tabular}[c]{@{}c@{}}NN(r=0.8,K=64)\\ MLP(256,128)\\ MLP(128, 128)\end{tabular}   & CLCA(256)                  & -                                                                                                &                            &                                                                                                    \\
                                \hline
\multirow{3}{*}{Segmentation}   & Path1                      & 512                            & \begin{tabular}[c]{@{}c@{}}NN(r=0.1, K=16)\\ MLP(3, 64)\\ MLP(64, 128)\\ MLP(128, 128)\end{tabular} & \begin{tabular}[c]{@{}c@{}}NN(r=0.2, K=32)\\ MLP(256,128)\\ MLP(128, 128)\end{tabular} & \begin{tabular}[c]{@{}c@{}}NN(r=0.4,, K=64)\\ MLP(256,128)\\ MLP(128, 128)\end{tabular} & CLCA(256)                  & \begin{tabular}[c]{@{}c@{}}UP(2048)\\ MLP(278, 512)\\ MLP(512, 256)\\ MLP(256, 128)\end{tabular} & \multirow{3}{*}{CSCA(128)} & \multirow{3}{*}{\begin{tabular}[c]{@{}l@{}}FC(128, 128)\\ FC(128, C)\end{tabular}}                 \\
                                & Path2                      & 256                            & \begin{tabular}[c]{@{}c@{}}NN(r=0.1, K=16)\\ MLP(3, 64)\\ MLP(64, 128)\\ MLP(128, 128)\end{tabular} & \begin{tabular}[c]{@{}c@{}}NN(r=0.2, K=32)\\ MLP(256,128)\\ MLP(128, 128)\end{tabular} & \begin{tabular}[c]{@{}c@{}}NN(r=0.4, K=64)\\ MLP(256,128)\\ MLP(128, 128)\end{tabular}  & CLCA(256)                  & \begin{tabular}[c]{@{}c@{}}UP(2048)\\ MLP(278, 512)\\ MLP(512, 256)\\ MLP(256, 128)\end{tabular} &                            &                                                                                                    \\
                                & Path3                      & 128                            & \begin{tabular}[c]{@{}c@{}}NN(r=0.1, K=16)\\ MLP(3, 64)\\ MLP(64, 128)\\ MLP(128, 128)\end{tabular} & \begin{tabular}[c]{@{}c@{}}NN(r=0.2, K=32)\\ MLP(256,128)\\ MLP(128, 128)\end{tabular} & \begin{tabular}[c]{@{}c@{}}NN(r=0.4,K=64)\\ MLP(256,128)\\ MLP(128, 128)\end{tabular}   & CLCA(256)                  & \begin{tabular}[c]{@{}c@{}}UP(2048)\\ MLP(278, 512)\\ MLP(512, 256)\\ MLP(256, 128)\end{tabular} &                            &                                                            \\
                                \hline
\end{tabular}}
\end{table*}

\section{Experiments}
We evaluate our CLCSCANet architecture with extensive experiments on multiple challenging benchmark datasets, including ModelNet \cite{wu20153d} for classification, ShapNetPart \cite{yi2016scalable} for part segmentation. And for all experiments, we use Pytorch framework to implement our proposed model on a NVIDIA RTX TITAN 24G GPU. Adam optimizer and step learning rate decay are adopted to train the network in end-to-end manner.


\subsection{3D Shape Classification}
ModelNet40 classification benchmark includes 12,311 meshed CAD models in 40 different object categories, where we randomly choose 9,843 models for training and 2,468 samples for evaluation. Following PointNet++ \cite{qi2017pointnet++}, 1,024 points without normals are uniformly sample from each object instance. Meanwhile, for fair comparison, data preprocessing operations, including random point dropout, random shifting, and random scale, are applied to augment the input. During training, we set the initial learning rate to 0.001, which is decayed 0.7 every 20 epochs. The classification network is trained for 150 epochs with a batch size of 20.

In Table \ref{tab:classification}, we report experimental settings and the quantitative comparison results with several state-of-the-art methods. From the table, it can be explicitly stated that 1) our CLCSCANet achieves comparable classification accuracy (92.2\%) among these state-of-the-art methods that only takes point cloud as input, such as PointNet \cite{qi2017pointnet}, OctreeGCNN \cite{lei2019octree} and SPH3D-GCN \cite{lei2020spherical}, etc. 2) Our model is slightly better than the PointNet++ with 5,000 points and normals as input, obtaining 0.3\% improvement, while it is a little worse than SFCNN \cite{rao2019spherical} that also uses normal as additional input. 3)Meanwhile, our CLCSCANet performs favorably against volumetric-based, like OctNet \cite{riegler2017octnet} and multi-view based models (e.g., MVCNN \cite{su2015multi}). 4) Experimental results show the effectiveness of our model in 3D object classification.

\begin{table}[]
\centering
\caption{3D Object Classification results on ModelNet40 dataset.}
\begin{tabular}{cccc} \hline
Method & Representation   & Input Size    & ModelNet40 \\ \hline
3DShapeNets \cite{wu20153d} & Volumetric  & $30^{3}$ & 77.3\% \\
VoxNet \cite{maturana2015voxnet} &  Volumetric & $32^{3}$   & 83.0\% \\
OctNet \cite{riegler2017octnet}  & Volumetric & $128^{3}$ &  86.5\% \\
\hline
MVCNN \cite{su2015multi}  & Multi-view & $12\times224^{2}$  & 90.1\%  \\
\hline
DeepNet \cite{ravanbakhsh2016deep} & Points & $5000\times3$  & 90.0\% \\

Kd-Net \cite{klokov2017escape} & Points  & $2^{15} \times 3$ &  88.5\% \\
PointNet \cite{qi2017pointnet} & Points & $1024 \times 3$  & 89.2\%  \\
ECC \cite{simonovsky2017dynamic} & Points & 1000 $\times$3 &   83.2\% \\
KC-Net \cite{shen2018mining} & Points & $1024\times3$   & 91.0\% \\
FoldingNet \cite{yang2018foldingnet} & Points  & $2048\times3$ &  88.4\%  \\
OctreeGCNN \cite{lei2019octree} & Points &  $1024\times3$ & 92.0\%\\ 
3D-GCN \cite{lin2020convolution} & Points & $1024\times3$  & 92.1\%\\
SPH3D-GCN \cite{lei2020spherical} & Points & $1000 \times 3$ &  92.1\% \\
\textbf{CLCSCANet} & Points & $1024 \times 3$ & \textbf{92.2}\%\\
\hline
SFCNN \cite{rao2019spherical} & Points+normals & $1024\times6$ &  92.3\% \\
PointNet++ \cite{qi2017pointnet++}& Points+normals & $5000\times6$ & 91.9\%   \\
\hline
\end{tabular}

\label{tab:classification}
\end{table}

\subsection{Part Segmentation}
For 3D point cloud segmentation task, we use ShapeNetPart \cite{yi2016scalable} dataset to validate the effectiveness of our CLCSCANet. This dataset consists of 16,881 3D CAD shapes from 16 different categories with 50 parts in total. We use the official split 14,007 shapes for training and 2,874 instances for testing, where each shape have 2 to 6 part labels, and each point is associated with a part label for segmentation task. Following the previous works, we adopt the average Intersection over Union (mIoU) over all instances and category-wise IoU over shapes under that category to evaluate the performance of our method. For fair comparison, we sample 2,048 points from each shape model. The initial learning rate is 0.0005 and is decayed half every 20 epochs. We use batch size of 8 to train the our segmentation network for 120 epochs. The momentum is 0.9.

\begin{figure*}[h]
  \centering
  \includegraphics[width=7in]{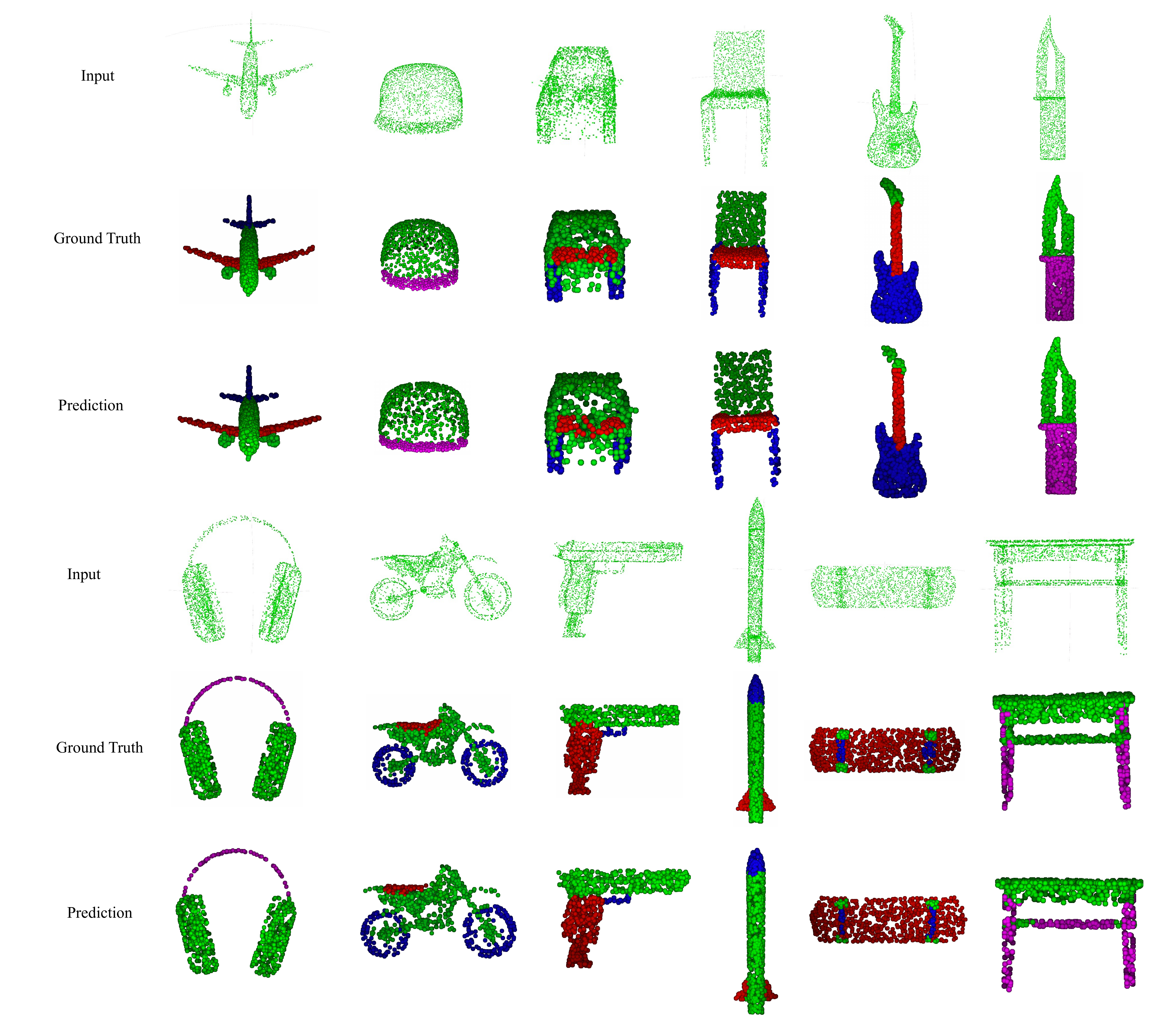}
  \caption{Qualitative comparison between the prediction of our CLCSCANet and the ground truth on ShapeNet Part dataset.}
  \label{fig_rsultforpart}
\end{figure*}

The 3D point cloud part segmentation results of our CLCSCANet are presented in Table \ref{PartSegmentation}, in which we also make an comparison with several state-of-the-art methods, such as SO-Net \cite{li2018so}, DGCNN \cite{wang2018dynamic}, 3D-GCN \cite{lin2020convolution}, etc. From our experimental results, we can report that our model achieves the highest mIoU of 85.3\%, and performs much better in 4 of 16 categories. Figure \ref{fig_rsultforpart} visualizes the qualitative comparison between the part segmentation results of our CLCSCANet and the ground truth. Both quantitative and visualization results demonstrate the success of our CLCSCANet in point cloud segmentation task.  

\begin{table*}[!t]
\caption{Experimental comparison of part segmentation with the state-of-the-art approaches on ShapeNet part dataset. The mean IoU across all the shape instances and IoU for each category are reported. }
\label{PartSegmentation}
\centering
\resizebox{\textwidth}{!}{
\begin{tabular}{c|c|cccccccccccccccc}
\hline
Method & mIoU & aero & bag & cap & car & chair & ep & guitar & knife & lamp & laptop & motor & mug & pistol & rocket & skate & table\\
\hline
ShapeNet \cite{yi2016scalable} & 81.4 & 81.0 & 78.4 & 77.7 & 75.7 & 87.6 & 61.9 & \textbf{92.0} & 85.4 & 82.5 & 95.7 & 70.6 & 91.9 & \textbf{85.9} & 53.1 & 69.8 & 75.3 \\
PointNet \cite{qi2017pointnet} & 83.7 & 83.4 & 78.7 & 82.5 & 74.9 & 89.6 & 73.0 & 91.5 & 85.9 & 80.8 & 95.3 & 65.2 & 93.0 & 81.2 & 57.9 & 72.8 & 80.6 \\
PointNet++ \cite{qi2017pointnet++} & 85.1 & 82.4 & 79.0 & 87.7 & 77.3 & 90.8 & 71.8 & 91.0 & 85.9 & 83.7 & 95.3 & 71.6 & 94.1 & 81.3 & 58.7 & \textbf{76.4} & 82.6\\
KD-Net \cite{klokov2017escape} & 82.3 & 80.1 & 74.6 & 74.3 & 70.3 & 88.6 & 73.5 & 90.2 & 87.2 & 71.0 & 94.9 & 57.4 & 86.7 & 78.1 & 51.8 & 69.9 & 80.3 \\
SO-Net \cite{li2018so} & 84.9 & 82.8 & 77.8 & 88.0 & 77.3 & 90.6 & 73.5 & 90.7 & 83.9 & 82.8 & 94.8 & 69.1 & 94.2 & 80.9 & 53.1 & 72.9 & 83.0 \\
RGCNN \cite{te2018rgcnn} & 84.3 & 80.2 & 82.8 & \textbf{92.6} & 75.3 & 89.2 & 73.7 & 91.3 & \textbf{88.4} & 83.3 & \textbf{96.0} & 63.9 & \textbf{95.7} & 60.9 & 44.6 & 72.9 & 80.4 \\
DGCNN \cite{wang2018dynamic} & 85.2 & 84.0 & 83.4 & 86.7 & 77.8 & 90.6 & 74.7 & 91.2 & 87.5 & 82.8 & 95.7 & 66.3 & 94.9 & 81.1 & \textbf{63.5} & 74.5 & 82.6\\
3D-GCN \cite{lin2020convolution} & 85.1 & 83.1 & \textbf{84.0} & 86.6 & 77.5 & 90.3 & 74.1 & 90.9 & 86.4 & 83.8 & 95.6 & 66.8 & 94.8 & 81.3 & 59.6 & 75.7 & 82.6 \\ 
ELM \cite{fujiwara2020neural} & 85.2 & \textbf{84.0} & 80.4 & 88.0 & \textbf{80.2} & 90.7 & \textbf{77.5} & 91.2 & 86.4 & 82.6 & 95.5 & 70.0 & 93.9 & 84.1 & 55.6 & 75.6 & 82.1 \\ 
Weak Sup. \cite{xu2020weakly} & 85.0 & 83.1 & 82.6 & 80.8 & 77.7 & 90.4 & 77.3 & 90.9 & 87.6 & 82.9& 95.8 & 64.7 & 93.9 &79.8 & 61.9& 74.9 & 82.9 \\
\hline
CLCSCANet & \textbf{85.3} & 82.7 & 71.3 & 77.8 & 77.8& \textbf{90.8} & 75.2 & 91.0 & 87.0 & \textbf{84.7} & 95.5 & \textbf{71.6} & 93.3 & 82.8 & 56.9 & 74.1 & \textbf{83.5} \\

\hline
\end{tabular}}
\end{table*}

\subsection{Ablation Study and Analysis} 
To validate the importance of our proposed modules in CLCSCANet, ablation studies are conducted on the classification task of ModelNet40. Here, we adopt a single-scale point-wise network that stacks multiple shared MLPs as our baseline. Table \ref{tab:ablation} reports the accuracy average class (ACC) and overall accuracy (OA) of different design choices of CLCSCANet. 

\begin{table}[]
\centering
\caption{Ablation studies on the ModelNet40 dataset. We analyze the effects of point-wise self-attention (PWSA) and channel-wise self-attention (CWSA).}
\begin{tabular}{ccc} \hline
Method & ACC  & OA  \\ \hline
Baseline & 85.1 & 87.1 \\
Baseline + CLCA & 89.5 & 91.6 \\
Baseline + CSCA & 88.7 & 90.8 \\
CLCSCANet &  \textbf{90.3}\% &  \textbf{92.2}\%\\
\hline
\end{tabular}

\label{tab:ablation}
\end{table}

From the table, we can see that both CLCA and CLCS modules significantly improve the classification performance in terms of ACC and OA. This remarkable improvements further convincingly demonstrate that the introduction of CLCA and CLCS modules enhance the geometric and semantic power of representation by modeling point-wise feature interactions across different levels and scales.

\section{Conclusion}

Following the remarkable advances in deep learning and self attention techniques, we propose a Cross-Level Cross-Scale Cross-Attention network architecture for better analyzing and understanding 3D point clouds in this paper. The introduction of Pointwise Feature Pyramid, Cross-Level Cross-Attention and Cross-Scale Cross-Attention models contributes to progressively aggregating geometric and semantic information and enhancing the representation capability. Results from extensive experiments show the effectiveness and competitive performance of our network. 



\bibliographystyle{ACM-Reference-Format}
\bibliography{sample-base}

\appendix

\end{document}